\pdfoutput=1

\documentclass[11pt]{article}

\usepackage[]{ACL2023}

\usepackage{times}
\usepackage{latexsym}

\usepackage[T1]{fontenc}

\usepackage[utf8]{inputenc}

\usepackage{microtype}

\usepackage{inconsolata}

\usepackage{hyperref}
\usepackage{algorithm}
\usepackage{algorithmic}
\usepackage{graphicx}
\usepackage{textcomp}
\usepackage{color}
\usepackage{xcolor}
\usepackage[utf8]{inputenc}
\usepackage{array}
\usepackage{pifont}
\usepackage{tabularx}
\usepackage{adjustbox}
\usepackage{booktabs,dcolumn}
\usepackage{makecell}
\usepackage{multirow}
\usepackage{multicol}
\usepackage{tablefootnote}
\usepackage{threeparttable}
\usepackage{float}
\usepackage{subfigure}
\usepackage{paralist}
\usepackage{comment}
\usepackage{enumerate}
\usepackage{xspace}
\usepackage{tcolorbox}
\usepackage{mathtools}
\usepackage{amsmath,amsthm,amsfonts,amssymb,bm,stmaryrd}
\usepackage{soul}
\usepackage{url}
\usepackage{fmtcount}
\usepackage{flushend}
\usepackage{thmtools}
\usepackage{colortbl}
\definecolor{mygray}{gray}{.9}

\usepackage{balance}

\usepackage{CJKutf8}

\newcommand{\eg}{\hbox{\emph{e.g.}}\xspace}

\newcommand{\ie}{\hbox{\emph{i.e.}}\xspace}

\DeclareMathOperator*{\argmin}{arg\,min}

\newcommand{\ours}{\textsc{Speech}}
\newcommand{\task}{ECSP}

\title{
\protect \includegraphics[width=0.028\linewidth]{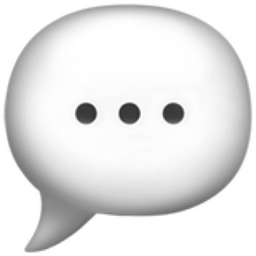} 
\textsc{Speech}: Structured Prediction with Energy-Based \\ Event-Centric Hyperspheres
}

\author{
  Shumin Deng$^{\heartsuit}$, Shengyu Mao$^{\spadesuit}$, Ningyu Zhang$^{\spadesuit}$\thanks{$\quad$ Corresponding Author.}~, Bryan Hooi$^{\heartsuit*}$ \\
  $^\heartsuit$National University of Singapore \& NUS-NCS Joint Lab, Singapore \\
  $^\spadesuit$Zhejiang University \& AZFT Joint Lab for Knowledge Engine, China \\
  \texttt{ 
    \{shumin,dcsbhk\}@nus.edu.sg, \{shengyu,zhangningyu\}@zju.edu.cn
  } \\
}

\begin{document}
\maketitle

\begin{abstract}
Event-centric structured prediction involves predicting structured outputs of events. In most NLP cases, event structures are complex with manifold dependency, and it is challenging to effectively represent these complicated structured events. To address these issues, we propose \textbf{S}tructured \textbf{P}rediction with \textbf{E}nergy-based \textbf{E}vent-\textbf{C}entric \textbf{H}yperspheres (\textbf{\ours}). 
{\ours} models complex dependency among event structured components with energy-based modeling, and represents event classes with simple but effective hyperspheres. 
Experiments on two unified-annotated event datasets indicate that {\ours} is predominant in event detection and event-relation extraction tasks. 

\end{abstract} 

\section{Introduction}
\label{sec:intro}

Structured prediction \cite{ICML2005_StructPred} is a task where the predicted outputs are complex structured components. This arises in many NLP tasks \cite{Book2011_LinguisticStructPred,ACL2017_BanditStructPred,ACL2023_Code4Struct} and supports various applications \cite{EMNLP2016_StructPred4Clinical,ACL-SPNLP2021_StructPred4Application}. 
In event-centric NLP tasks, there exists strong complex dependency between the structured outputs, such as event detection (ED) \cite{ACL2015_DMCNN}, event-relation extraction (ERE) \cite{J2020_EE_ERE}, and event schema induction \cite{EMNLP2020_PathLM}. Thus, these tasks can also be revisited as event-centric structured prediction problems \cite{ACL2013_EE-StructPred}. 

\begin{figure}[t] 
  \centering
  \includegraphics[width=0.96\linewidth]{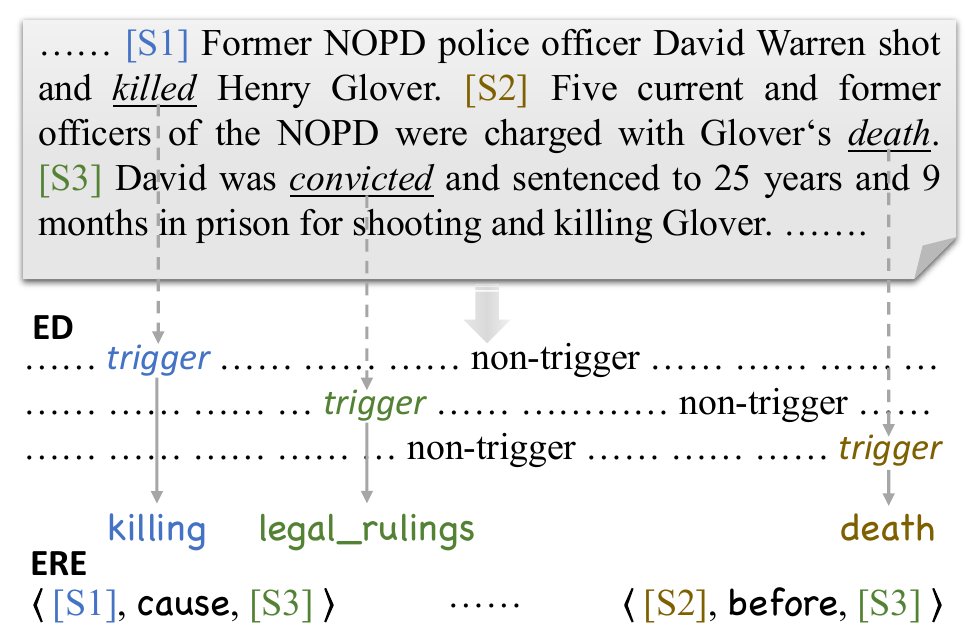}
  \vspace{-2mm}
  \caption{Illustration of event-centric structured prediction tasks, with the examples of ED and ERE. 
  \label{fig:intro_task} }
  \vspace{-3mm}
\end{figure}

Event-centric structured prediction ({\task}) tasks require to consider manifold structures and dependency of events, including intra-/inter-sentence structures. For example, as seen in Figure~\ref{fig:intro_task}, given a document containing some event mentions ‘‘\emph{David Warren shot and killed Henry Glover ... David was convicted and sentenced to 25 years and 9 months ...}'', in ED task mainly considering intra-sentence structures, we need to identify event triggers (\emph{killed}, \emph{convicted}) from these tokens and categorize them into event classes (\emph{killing}, \emph{legal\_rulings}); in ERE task mainly considering inter-sentence structures, we need to find the relationship between each event mention pair, such as event coreference, temporal, causal and subevent relations. 

As seen from Figure~\ref{fig:intro_task}, the outputs of {\task} lie on a complex manifold and possess interdependent structures, \eg, the long-range dependency of tokens, the association among triggers and event classes, and the dependency among event classes and event relations. Thus it is challenging to model such complex event structures while efficiently representing these events. 
Previous works increasingly apply deep representation learning to tackle these problems. 
\citet{ACL2020_OneIE,EMNLP2020_PathLM} propose to predict event structures based on the event graph schema. 
\citet{NAACL2022_DEGREE} generate event structures with manually designed prompts. 
However, these methods mainly focus on one of {\task} tasks and their event structures are hard to represent effectively. 
\citet{ICLR2021_TANL,ACL2021_Text2Event,ACL2022_UIE} propose to extract multiple event structures from texts with a unified generation paradigm. 
However, the event structures of these approaches are usually quite simplistic and they often ignore the complex dependency among tasks. In this paper, we focus more on: (i) how to learn complex event structures for manifold {\task} tasks; and (ii) how to simultaneously represent events for these complex structured prediction models effectively. 

To resolve the first challenging problem of modeling manifold event structures, we utilize energy networks \cite{Tutorial2006_EnergyBasedLearn,ICML2016_EneNet4StructPred,ICML2017_EneNet4StructPred,ICLR2018_EneNet4StructPred}, inspired by their potential benefits in capturing complex dependency of structured components. 
We define the energy function to evaluate compatibility of input/output pairs, which places no limits on the size of the structured components, making it powerful to model complex and manifold event structures. 
We generally consider token-, sentence-, and document- level energy respectively for trigger classification, event classification and event-relation extraction tasks. 
To the best of our knowledge, this work firstly address event-centric structured prediction with energy-based modeling. 

To resolve the second challenging problem of efficiently representing events, we take advantage of hyperspheres \cite{NIPS2019_HPN,ICML2020_HyperSphere}, which is demonstrated to be a simple and effective approach to model class representation \cite{KBS2022_kHPN}. 
We assume that the event mentions of each event class distribute on the corresponding energy-based hypersphere, so that we can represent each event class with a hyperspherical centroid and radius embedding. The geometrical modeling strategy \cite{ICLR2021_PN4RE,EMNLP2021_PN4ED} is demonstrated to be beneficial for modelling enriched class-level information and suitable for constructing measurements in Euclidean space, making it intuitively applicable to manifold event-centric structured prediction tasks. 

Summarily, considering the two issues, we propose to address \textbf{S}tructured \textbf{P}rediction with \textbf{E}nergy-based \textbf{E}vent-\textbf{C}entric \textbf{H}yperspheres (\textbf{\ours}), and 
our contributions can be summarized as follows: 

\begin{itemize}

	\item We revisit the event-centric structured prediction tasks in consideration of both complex event structures with manifold dependency and efficient representation of  events. 

	\item We propose a novel approach named {\ours} to model complex event structures with energy-based networks and efficiently represent events with event-centric hyperspheres. 

	\item We evaluate {\ours} on two newly proposed datasets for both event detection and event-relation extraction, and experiments demonstrate that our model is advantageous. 

\end{itemize}

\section{Related Work}
\label{sec:related_work}

\textbf{Event-Centric Structured Prediction (\task). }
Since the boom in deep learning, traditional approaches to {\task} mostly \emph{define a score function between inputs and outputs based on a neural network}, such as CNN \cite{ACL2015_DMCNN,WSDM2020_MetaL-EE_DMBPN}, RNN \cite{NAACL2016_JRNN,EMNLP2018_ERE-T,AAAI2019_Joint3EE}, and GCN \cite{EMNLP2019_MOGANED,EMNLP2020_GatedGCN,EMNLP2020_EE-GCN}. 
With the development of pretrained large models, more recent research has entered a new era. \citet{EMNLP2019_AD-DMBERT,EMNLP2020_QAEE,EMNLP2020_RCEE,ACL2021_OntoED,SIGIR2022_CorED} leverage BERT \cite{NAACL2019_BERT} for event extraction. 
\citet{EMNLP2020_ERE_T} and \citet{EMNLP2020_ERE_H-T,AAAI2022_ERE,ACL2022_ERE} respectively adopt BERT and RoBERTa \cite{arXiv2019_RoBERTa} for event-relation extraction. 
\citet{ACL2021_Text2Event,ICLR2021_TANL,ACL2022_UIE} propose generative {\task} models based on pre-trained T5 \cite{J2020_T5}. 
\citet{ACL2023_Code4Struct} tackle {\task} with code generation based on code pretraining. 
However, these approaches are equipped with fairly simplistic event structures and have difficulty in tackling complex dependency in events. Besides, most of them fail to represent manifold events effectively. 

\textbf{Energy Networks for Structured Prediction and Hyperspheres for Class Representation. }
Energy networks \emph{define an energy function over input/output pairs with arbitrary neural networks}, which places no limits on the size of the structured components, making it advantageous in modeling complex and manifold event structures. 
\citet{Tutorial2006_EnergyBasedLearn,ICML2016_EneNet4StructPred} associate a scalar measure to evaluate the compatibility to each configuration of inputs and outputs. \cite{ICML2016_EneNet4StructPred} formulate deep energy-based models for structured prediction, called structured prediction energy networks (SPENs). \citet{ICML2017_EneNet4StructPred} present end-to-end learning for SPENs, \citet{ICLR2018_EneNet4StructPred} jointly train structured energy functions and inference networks with large-margin objectives. 
Some previous researches also regard event-centric NLP tasks as structured prediction \cite{ACL2013_EE-StructPred,ICLR2021_TANL}. 
Furthermore, to effectively obtain event representations, \citet{KBS2022_kHPN} demonstrate that hyperspherical prototypical networks \cite{NIPS2019_HPN} are powerful to encode enriched semantics and dependency in event structures, but they merely consider support for pairwise event structures. 

\section{Methodology}
\label{sec:method}

\subsection{Preliminaries}
For structured prediction tasks, given input $\bm{x} \in \mathcal{X}$, we denote the structured outputs by $\mathbf{M}_{\Phi}(\bm{x}) \in \tilde{\mathcal{Y}}$ with a prediction model $\mathbf{M}_{\Phi}$. 
Structured Prediction Energy Networks (SPENs) score structured outputs with an \textbf{energy function} $E_{\Theta} : \mathcal{X} \times \tilde{\mathcal{Y}} \rightarrow \mathbb{R}$ parameterized by $\Theta$ that iteratively optimize the energy between the input/output pair \cite{ICML2016_EneNet4StructPred}, where lower energy means greater compatibility between the pair. 

We introduce event-centric structured prediction (\task) following the similar setting as SPENs for multi-label classification and sequence labeling proposed by \citet{ICLR2018_EneNet4StructPred}. Given a feature vector $\bm{x}$ belonging to one of $T$ labels, the model output is $\mathbf{M}_{\Phi}(\bm{x}) = \{0, 1\}^T \in \tilde{\mathcal{Y}}$ for all $\bm{x}$. 
The energy function contains two terms: 
\begin{equation}
\begin{aligned}
E_{\Theta}(\bm{x}, \bm{y}) & = E^{local}_{\Theta}(\bm{x}, \bm{y})  + E^{label}_{\Theta}(\bm{y}) \\ 
& = \sum_{i=1}^T y_{i} V_{i}^\top f(\boldsymbol{x}) + w^\top g(W \bm{y})
\label{eq:energy_function}
\end{aligned}
\end{equation}
where $E^{local}_{\Theta}(\bm{x}, \bm{y}) = \sum_{i=1}^T y_{i} V_{i}^\top f(\boldsymbol{x})$ is the sum of linear models, and 
$y_{i} \in \bm{y}$, $V_{i}$ is a parameter vector for label $i$ and $f(\bm{x})$ is a multi-layer perceptron computing a feature representation for the input $\bm{x}$; 
$E^{label}_{\Theta}(\bm{y}) = w^\top g(W \bm{y})$ returns a scalar which quantifies the full set of labels, scoring $\bm{y}$ independent of $\bm{x}$, thereinto, 
$w$ is a parameter vector, $g(\cdot)$ is an elementwise non-linearity function, and $W$ is a parameter matrix learned from data indicating the interaction between labels. 

After learning the energy function, prediction minimizes energy:
\begin{equation}
\tilde{\bm{y}} = \argmin_{\bm{y} \in \tilde{\mathcal{Y}}} E_{\Theta}(\bm{x}, \bm{y})
\label{eq:spen_inf}
\end{equation}

The final theoretical optimum for SPEN is denoted by: 
\begin{equation}
\begin{aligned}
\min_{\Theta} \max_{\Phi} \sum \big[ & \triangle \left(\mathbf{M}_{\Phi}(\bm{x}_i), \bm{y}_i\right) - \\
& E_{\Theta}\left(\bm{x}_i, \mathbf{M}_{\Phi}(\bm{x}_i)\right) + E_{\Theta}\left(\bm{x}_i, \bm{y}_i\right) \big]_{+} 
\label{eq:spen_loss}
\end{aligned}
\end{equation}
where $[a]_{+} = \max(0, a)$, and $\triangle(\tilde{\boldsymbol{y}}, \boldsymbol{y})$, often referred to ‘‘margin-rescaled'' structured hinge loss, is a structured cost function that returns a nonnegative value indicating the difference between the predicted result $\tilde{\boldsymbol{y}}$ and ground truth $\boldsymbol{y}$. 


\subsection{Problem Formulation}
In this paper, we focus on {\task} tasks of event detection (ED) and event-relation extraction (ERE). 
ED can be divided into trigger classification for tokens and event classification for sentences. 
We denote the dataset by $\mathcal{D} = \{ \mathcal{E}, \mathcal{R}, \mathcal{X} \}$ containing an event class set $\mathcal{E}$, a multi-faceted event-relation set $\mathcal{R}$ and the event corpus $\mathcal{X}$, thereinto, 
$\mathcal{E} = \{ e_i \ | \ i \in [1, |\mathcal{E}|] \}$ contains $|\mathcal{E}|$ event classes including a None; 
$\mathcal{R} = \{ r_i \ | \ i \in [1, |\mathcal{R}|] \}$ contains  $|\mathcal{R}|$ temporal, causal, subevent and coreference relationships among event mentions including a NA event-relation; 
$\mathcal{X} = \{ \boldsymbol{X}_i \ | \ i \in [1, K] \}$ consists of $K$ event mentions, where $\boldsymbol{X}_i$ is denoted as a token sequence $\boldsymbol{x} = \{\boldsymbol{x}_j \ | \ j \in [1, L]\}$ with maximum $L$ tokens. 
For \emph{trigger classification}, the goal is to predict the index $t$ ($1 \leq t \leq L$) of the trigger $\boldsymbol{x}_t$ in each token sequence $\boldsymbol{x}$ and categorize $\boldsymbol{x}_t$ into a specific event class $e_i \in \mathcal{E}$. 
For \emph{event classification}, we expect to predict the event label $e_i$ for each event mention $\boldsymbol{X}_i$. 
For \emph{event-relation extraction}, we require to identify the relation $r_i \in \mathcal{R}$ for a pair of event mentions $\ddot{\boldsymbol{X}}_{\langle ij \rangle} = ( \boldsymbol{X}_i, \boldsymbol{X}_j )$. 

In summary, our goal is to design an {\task} model $\mathbf{M}_{\Phi}$, aiming to tackle the tasks of: 
(1) \emph{trigger classification}: to predict the token label $\tilde{\boldsymbol{y}} = \mathbf{M}_{\Phi}(\boldsymbol{x})$ for the token list $\boldsymbol{x}$; 
(2) \emph{event classification}: to predict the event class label $\tilde{\boldsymbol{Y}} = \mathbf{M}_{\Phi}(\boldsymbol{X})$ for the event mention $\boldsymbol{X}$;
(3) \emph{event-relation extraction}: to predict the event-relation label $\tilde{\boldsymbol{z}} = \mathbf{M}_{\Phi}(\ddot{\boldsymbol{X}})$ for the event mention pair $\ddot{\boldsymbol{X}}$.

\subsection{Model Overview}

\begin{figure*}[!htbp]
  \centering
  \includegraphics[width=0.92\linewidth]{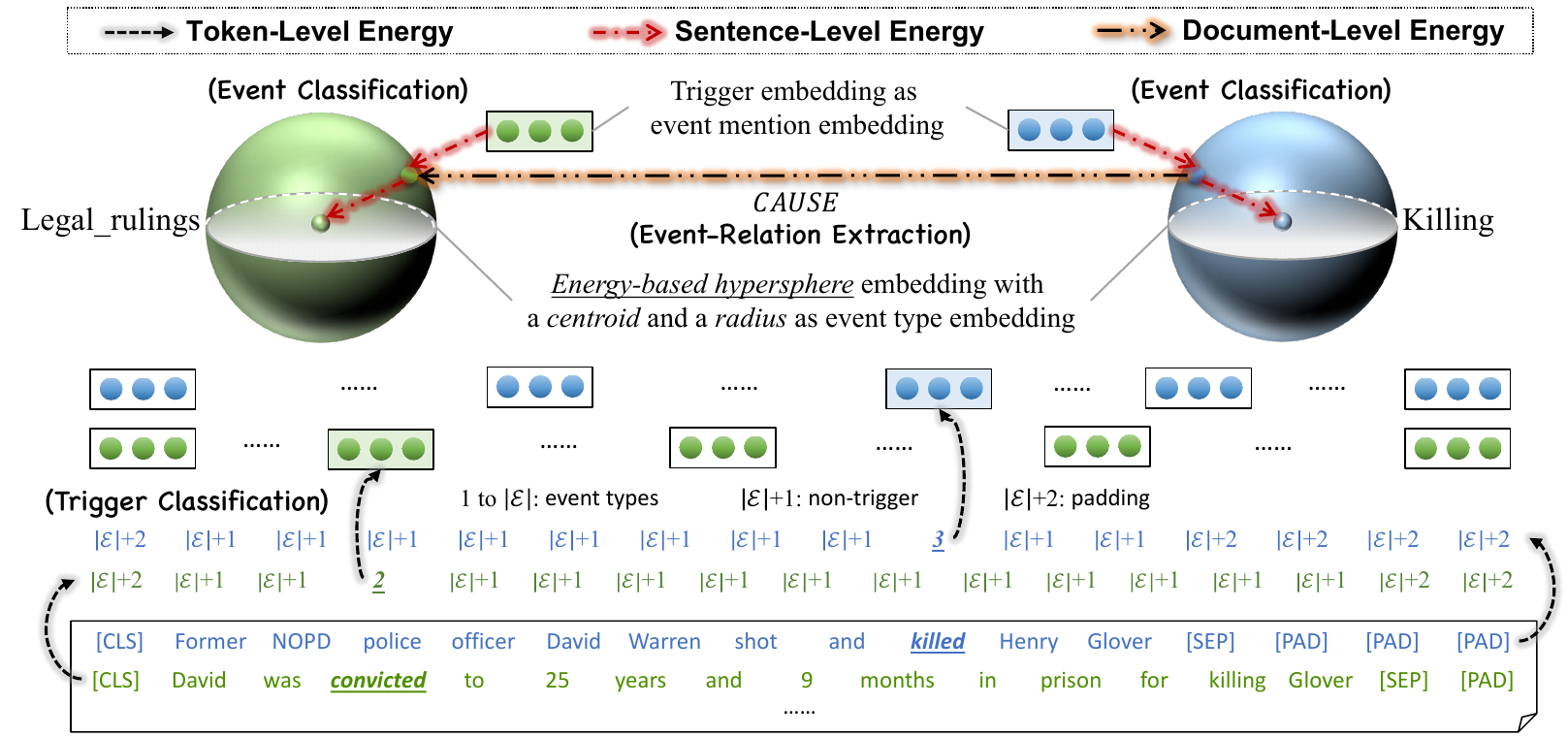}
  \vspace{-3mm}
  \caption{Overview of {\ours} with examples, where 
  token-level energy serves for event trigger classification, 
  sentence-level energy serves for event classification and 
  document-level energy serves for event-relation extraction. 
  \label{fig:model_overview} }
  \vspace{-6mm}
\end{figure*}

As seen in Figure~\ref{fig:model_overview}, {\ours} combines three levels of energy: token, sentence, as well as document, and they respectively serve for three kinds of {\task} tasks: 
(1) token-level energy for trigger classification: considering energy-based modeling is able to capture long-range dependency among tokens without limits to token size; 
(2) sentence-level energy for event classification: considering energy-based hyperspheres can model the complex event structures and represent events efficiently; and 
(3) document-level energy for event-relation extraction: considering energy-based modeling enables us to address the association among event mention pairs and event-relations. 
We leverage the trigger embeddings as event mention embeddings; the energy-based hyperspheres with a centroid and a radius as event class embeddings, and these three tasks are associative to each other. 

\subsection{Token-Level Energy}

Token-level energy serves for trigger classification. 
Given a token sequence $\boldsymbol{x} = \{\boldsymbol{x}_j | j \in [1, L]\}$ with trigger $\boldsymbol{x}_t$, we leverage a pluggable backbone encoder to obtain the contextual representation $f_1(\boldsymbol{x})$ for each token, such as pre-trained BERT \cite{NAACL2019_BERT}, RoBERTa \cite{arXiv2019_RoBERTa}, DistilBERT \cite{NeurIPS2019-EMC2_DistilBERT} and so on. 
We then predict the label $\tilde{\boldsymbol{y}} = \mathbf{M}_{\Phi}(\boldsymbol{x})$ of each token with an additional linear classifier. Inspired by SPENs for sequence labeling \cite{ICLR2018_EneNet4StructPred}, we also adopt an energy function for token classification. 

\textbf{Energy Function.} The token-level energy function is inherited from Eq~\eqref{eq:energy_function}, defined as: 
\begin{equation}
\begin{aligned}
& E_{\Theta}(\boldsymbol{x}, \boldsymbol{y}) = \\ 
& - \left( 
 \sum_{n=1}^L \sum_{i=1}^{|\mathcal{E}|+2} \underbrace{ y_n^i \left( V_{1,i}^{\top} f_1(\boldsymbol{x}_n) \right) }_{local}
 + \sum_{n=1}^{L} \underbrace{ y_{n-1}^{\top} W_1 y_n }_{label}
 \right)
\label{eq:token_energy}
\end{aligned}
\end{equation}
where 
$y_n^i$ is the $i_{th}$ entry of the vector $y_n \in \boldsymbol{y}$, indicating the probability of the $n_{th}$ token $\boldsymbol{x}_n$ being labeled with $i$ ($i$ for $e_i$, $|\mathcal{E}|$+1 for non-trigger and $|\mathcal{E}|$+2 for padding token). $f_1(\cdot)$ denotes the feature encoder of tokens. Here our learnable parameters are $\Theta = (V_1, W_1)$, thereinto, 
$V_{1,i} \in \mathbb{R}^d$ is a parameter vector for token label $i$, and
$W_1 \in \mathbb{R}^{(|\mathcal{E}|+2) \times (|\mathcal{E}|+2)}$ contains the bilinear product between $y_{n-1}$ and $y_n$ for token label pair terms. 

\textbf{Loss Function.} 
The training objective for trigger classification is denoted by: 
\begin{equation}
\begin{aligned}
\mathcal{L}_{tok} = &
\sum\nolimits_{i=1}^{L} \big[ \triangle \left( \tilde{\boldsymbol{y}}_i, \boldsymbol{y}_i \right) - E_{\Theta} \left( \boldsymbol{x}_i, \tilde{\boldsymbol{y}}_i \right) \\ 
& + E_{\Theta} \left( \boldsymbol{x}_i, \boldsymbol{y}_i \right) \big]_{+} 
+ \mu_1 \mathcal{L}_{\mathrm{CE}}\left( \tilde{\boldsymbol{y}}_i, \boldsymbol{y}_i \right)
\label{eq:loss_token}
\end{aligned}
\end{equation}
where $\tilde{\boldsymbol{y}}_i$ and $\boldsymbol{y}_i$ respectively denote predicted results and ground truth. The first half of Eq~\eqref{eq:loss_token} is inherited from Eq~\eqref{eq:spen_loss} for the energy function, and in the latter half,  
$\mathcal{L}_{\mathrm{CE}}\left( \tilde{\boldsymbol{y}}_i, \boldsymbol{y}_i \right)$ is the trigger classification cross entropy loss, and $\mu_1$ is its ratio. 

\subsection{Sentence-Level Energy}

Sentence-level energy serves for event classification. 
Given the event mention $\boldsymbol{X}_i$ with the trigger $\boldsymbol{x}_t$, we utilize the trigger embedding $f_1(\boldsymbol{x}_t)$ as the event mention embedding $f_2(\boldsymbol{X})$, where $f_2(\cdot)$ denotes the feature encoder of event mentions. 
We then predict the class of each event mention with energy-based hyperspheres, denoted by $\tilde{\boldsymbol{Y}} = \mathbf{M}_{\Phi}(\boldsymbol{X})$. 

Specifically, we use an energy-based hypersphere to represent each event class, and assume that the event mentions of each event class should distribute on the corresponding hypersphere with the lowest energy. We then calculate the probability of the event mention $\boldsymbol{X}$ categorizing into the class $e_i$ with a \textbf{hyperspherical measurement function}: 
\vspace{-2mm}
\begin{equation}
\mathcal{S}(\boldsymbol{X}, \mathcal{P}_i) 
= \frac{ \mathrm{exp}^{- [ ~ \| \mathcal{P}_i - f_2(\boldsymbol{X}) \|_2 - \gamma ~ ]_{+}} }{ \sum\nolimits_{j=1}^{|\mathcal{E}|} \mathrm{exp}^{- [ ~ \| \mathcal{P}_j - f_2(\boldsymbol{X}) \|_2 - \gamma ~ ]_{+}} }
\label{eq:distance}
\end{equation}
where $[a]_{+} = \max(0, a)$, 
$\mathcal{P}_i$ denotes the hypersphere centroid embedding of $e_i$. 
$\|\cdot\|$ denotes the Euclidean distance. 
$\gamma$ is the radius of the hypersphere, which can be scalable or constant. We simply set $\gamma = 1$ in this paper, meaning that each event class is represented by a unit hypersphere. Larger $\mathcal{S}(\boldsymbol{X}, \mathcal{P}_i)$ signifies that the event mention $\boldsymbol{X}$ are more likely be categorized into $\mathcal{P}_i$ corresponding to $e_i$. 
To measure the energy score between event classes and event mentions, we also adopt an energy function for event classification. 

\textbf{Energy Function.} The sentence-level energy function is inherited from Eq~\eqref{eq:energy_function}, defined as:  
\begin{equation}
\begin{aligned}
& E_{\Theta}(\boldsymbol{X}, \boldsymbol{Y}) = \\
& - \left( 
 \sum_{i=1}^{|\mathcal{E}|} \underbrace{ \boldsymbol{Y}_i \left(V_{2,i}^{\top} f_2(\boldsymbol{X})\right) }_{local}
 + \underbrace{ w_2^{\top} g(W_2 \boldsymbol{Y}) }_{label}
 \right)
\end{aligned}
\label{eq:sent_energy}
\end{equation}
where 
$\boldsymbol{Y}_i \in \boldsymbol{Y}$ indicates the probability of the event mention $\boldsymbol{X}$ being categorized to $e_i$. 
Here our learnable parameters are $\Theta = (V_2, w_2, W_2)$, thereinto,  
$V_{2,i} \in \mathbb{R}^d$ is a parameter vector for $e_i$,  
$w_2 \in \mathbb{R}^{|\mathcal{E}|}$ and $W_2 \in \mathbb{R}^{|\mathcal{E}| \times |\mathcal{E}|}$. 

\textbf{Loss Function.}
The training objective for event classification is denoted by: 
\begin{equation}
\begin{aligned}
\mathcal{L}_{sen} = & \sum\nolimits_{i=1}^K 
\big[ \triangle \left( \tilde{\boldsymbol{Y}}_i, \boldsymbol{Y}_i \right) 
- E_{\Theta} \left( \boldsymbol{X}_i, \tilde{\boldsymbol{Y}}_i \right) \\
& + E_{\Theta} \left( \boldsymbol{X}_i, \boldsymbol{Y}_i \right)
\big]_{+} 
+ \mu_2 \mathcal{L}_{\mathrm{CE}}\left( \tilde{\boldsymbol{Y}}_i, \boldsymbol{Y}_i \right)
\label{eq:loss_sent}
\end{aligned}
\end{equation}
where the first half is inherited from Eq~\eqref{eq:spen_loss}, and in the latter half, 
$\mathcal{L}_{\mathrm{CE}}$ is a cross entropy loss for predicted results $\tilde{\boldsymbol{Y}}_i$ and ground truth $\boldsymbol{Y}_i$. 
$\mu_2$ is a ratio for event classification cross entropy loss. 

\subsection{Document-Level Energy}
Document-level energy serves for event-relation extraction. 
Given event mentions $\boldsymbol{X}$ in each document, we model the embedding interactions of each event mention pair with a comprehensive feature vector $ f_3(\ddot{\boldsymbol{X}}_{\langle ij \rangle}) = \big[ f_2(\boldsymbol{X}_i), f_2(\boldsymbol{X}_j), f_2(\boldsymbol{X}_i) \odot f_2(\boldsymbol{X}_j) \big]$. 
We then predict the relation between each event mention pair with a linear classifier, denoted by $\tilde{\boldsymbol{z}} = \mathbf{M}_{\Phi}(\ddot{\boldsymbol{X}})$. Inspired by SPENs for multi-label classification \cite{ICLR2018_EneNet4StructPred}, we also adopt an energy function for ERE. 

\textbf{Energy Function.} The document-level energy function is inherited from Eq~\eqref{eq:energy_function}, defined as:  
\begin{equation}
\begin{aligned}
& E_{\Theta}(\ddot{\boldsymbol{X}}, \boldsymbol{z}) = \\ 
& - \left( 
 \sum_{i=1}^{|\mathcal{R}|} \underbrace{ \boldsymbol{z}_i \left( V_{3,i}^{\top} f_3(\ddot{\boldsymbol{X}}) \right) }_{local}
 + \underbrace{ w_3^{\top} g(W_3 \boldsymbol{z}) }_{label}
 \right)
\end{aligned}
\label{eq:doc_energy}
\end{equation}
where 
$\boldsymbol{z}_i \in \boldsymbol{z}$ indicates the probability of the event mention pair $\ddot{\boldsymbol{X}}$ having the relation of $r_i$. 
Here our learnable parameters are $\Theta = (V_3, w_3, W_3)$, thereinto, 
$V_{3,i} \in \mathbb{R}^{3d}$ is a parameter vector for $r_i$,  
$w_3 \in \mathbb{R}^{|\mathcal{R}|}$ and $W_3 \in \mathbb{R}^{|\mathcal{R}| \times |\mathcal{R}|}$. 

\textbf{Loss Function.} 
The training objective for event-relation extraction is denoted by: 
\begin{equation}
\begin{aligned}
\mathcal{L}_{doc} = & \sum\nolimits_{k=1}^N 
\Big[ \triangle \left( \tilde{\boldsymbol{z}}_k, \boldsymbol{z}_{k} \right) 
- E_{\Theta} \left( \ddot{\boldsymbol{X}}_{k}, \tilde{\boldsymbol{z}}_k \right) \\
& + E_{\Theta} \left( \ddot{\boldsymbol{X}}_{k}, \boldsymbol{z}_{k} \right)
\Big]_{+} 
+ \mu_3 \mathcal{L}_{\mathrm{CE}} \left( \tilde{\boldsymbol{z}}_k, \boldsymbol{z}_{k} \right)
\label{eq:loss_doc}
\end{aligned}
\end{equation}
where the first half is inherited from Eq~\eqref{eq:spen_loss}, and in the latter half, 
$\mathcal{L}_{\mathrm{CE}}\left( \tilde{\boldsymbol{z}}_k, \boldsymbol{z}_{k} \right)$ is the event-relation extraction cross entropy loss, $\mu_3$ is its ratio, and $N$ denotes the quantity of event mention pairs. 

The \textbf{final training loss} for {\ours} $\mathbf{M}_{\Phi}$ parameterized by $\Phi$ is defined as: 
\begin{equation}
\mathcal{L} = 
\lambda_1\mathcal{L}_{tok} + \lambda_2\mathcal{L}_{sen} + \lambda_3\mathcal{L}_{doc}
+ \| \Phi\|_2^2 
\label{eq:loss_all}
\end{equation}
where $\lambda_1$, $\lambda_2$, $\lambda_3$ are the loss ratios respectively for trigger classification, event classification and event-relation extraction tasks. We add the penalty term $\| \Phi\|_2^2$ with $L_2$ regularization.

\section{Experiments}
\label{sec:experiment}

The experiments refer to event-centric structured prediction (\task) and comprise three tasks: 
(1) Trigger Classification;
(2) Event Classification; and 
(3) Event-Relation Extraction.

\subsection{Datasets and Baselines}

\begin{table}[!htbp] 
\centering
\small
	\begin{tabular}{l | c | c }
	
	\toprule

	& \textbf{\textsc{Maven-Ere}} & \textbf{\textsc{OntoEvent-Doc}} \\

	\midrule

	\# Document & 4,480 & 4,115 \\

	\# Mention & 112,276 & 60,546 \\

	\# Temporal & 1,216,217 & 5,914 \\

	\# Causal & 57,992 & 14,155 \\

	\# Subevent & 15,841 & / \\
	
	\bottomrule
	\end{tabular}
	\caption{
	The statistics about \textsc{Maven-Ere} and \textsc{OntoEvent-Doc} used in this paper. 
	\label{tab:exp_data_stat}
	}
\end{table}

\begin{table*}[t] 
\centering
\small
	\begin{tabular}{l | c c c | c c c}
	
	\toprule

	\multirow{3}*{\textbf{Model}} & \multicolumn{3}{c|}{\textbf{\textsc{Maven-Ere}}} & \multicolumn{3}{c}{\textbf{\textsc{OntoEvent-Doc}}} \\

	\cmidrule{2-7}

	& \textbf{P} & \textbf{R} & \textbf{F1} & \textbf{P} & \textbf{R} & \textbf{F1} \\

	\midrule

	DMCNN$^{\dagger}$ 
    & 60.09 $\pm$ 0.36 & 60.34 $\pm$ 0.45 & 60.21 $\pm$ 0.21 
	& 50.42 $\pm$ 0.99 & 52.24 $\pm$ 0.46 & 51.31 $\pm$ 0.39 \\

	BiLSTM-CRF$^{\dagger}$ 
    & 61.30 $\pm$ 1.07 & 64.95 $\pm$ 1.03 & 63.06 $\pm$ 0.23 
	& 48.86 $\pm$ 0.81 & 55.91 $\pm$ 0.56 & 52.10 $\pm$ 0.43 \\

	DMBERT$^{\dagger}$ 
 	& 56.79 $\pm$ 0.54 & \underline{76.24} $\pm$ 0.26 & 65.09 $\pm$ 0.32 
	& 53.82 $\pm$ 1.01 & \underline{66.12} $\pm$ 1.02 & 59.32 $\pm$ 0.24 \\

	BERT-CRF$^{\dagger}$ 
 	& 62.79 $\pm$ 0.34 & 70.51 $\pm$ 0.94 & 65.73 $\pm$ 0.57
	& 52.18 $\pm$ 0.81 & 62.31 $\pm$ 0.45 & 56.80 $\pm$ 0.53 \\ 



	MLBiNet$^{\ddagger}$ 
	& 63.50 $\pm$ 0.57 & 63.80 $\pm$ 0.47 & 63.60 $\pm$ 0.52 
	& 56.09 $\pm$ 0.93 & 57.67 $\pm$ 0.81 & 56.87 $\pm$ 0.87 \\


	TANL$^{\ddagger}$ 
	& \underline{68.66} $\pm$ 0.18 & 63.79 $\pm$ 0.19 & 66.13 $\pm$ 0.15 
	& 57.73 $\pm$ 0.65 & 59.93 $\pm$ 0.31 & 59.13 $\pm$ 0.52 \\

	\textsc{Text2Event}$^{\ddagger}$ 
	& 59.91 $\pm$ 0.83 & 64.62 $\pm$ 0.65 & 62.16 $\pm$ 0.25 
	& 52.93 $\pm$ 0.94 & 62.27 $\pm$ 0.49 & 57.22 $\pm$ 0.75 \\



	CorED-BERT$^{\ddagger}$ 
	& 67.62 $\pm$ 1.03 & 69.49 $\pm$ 0.63 & \underline{68.49} $\pm$ 0.42 
	& \underline{60.27} $\pm$ 0.55 & 62.25 $\pm$ 0.66 & \underline{61.25} $\pm$ 0.19  \\

	\midrule

	\textbf{\textsc{\ours}} 
	& \textbf{78.82} $\pm$ 0.82 & \textbf{79.37} $\pm$ 0.75 & \textbf{79.09} $\pm$ 0.82 
	& \textbf{74.67} $\pm$ 0.58 & \textbf{74.73} $\pm$ 0.62 & \textbf{74.70} $\pm$ 0.58 \\

	\quad w/o energy
	& 76.12 $\pm$ 0.32 & 76.66 $\pm$ 0.25 & 76.38 $\pm$ 0.28  
	& 71.76 $\pm$ 0.38 & 72.17 $\pm$ 0.39 & 71.96 $\pm$ 0.38  \\


	\bottomrule
	\end{tabular}
	\caption{
	Performance (\%) of trigger classification on \textsc{Maven-Ere} \emph{valid set} and \textsc{OntoEvent-Doc} \emph{test set}. 
	$\dagger$: results are produced with codes referred to \citet{EMNLP2020_MAVEN};  
	$\ddagger$: results are produced with official implementation. 
	\textbf{Best results} are marked in bold, and the \underline{second best results}  are underlined. 
	\label{tab:exp_token_ed}
	}
\vspace{-2mm}

\end{table*}

\emph{Datasets.} 
Considering event-centric structured prediction tasks in this paper require fine-grained annotations for events, such as labels of tokens, event mentions, and event-relations, we select two newly-proposed datasets meeting the requirements: 
\textsc{Maven-Ere} \cite{EMNLP2022_MAVEN-ERE} and \textsc{OntoEvent-Doc} \cite{ACL2021_OntoED}. 
Note that \textsc{OntoEvent-Doc} is derived from \textsc{OntoEvent} \cite{ACL2021_OntoED} which is formatted in a sentence level. We reorganize it and make it format in a document level, similar to \textsc{Maven-Ere}. Thus the train, validation, and test sets of \textsc{OntoEvent-Doc} are also different from the original \textsc{OntoEvent}. 
We release the reconstructed dataset and code in Github\footnote{\url{https://github.com/zjunlp/SPEECH}.} for reproduction. 
To simplify the experiment settings, we dismiss hierarchical relations of \textsc{OntoEvent} and coreference relations of \textsc{Maven-Ere} in this paper. More details of multi-faceted event-relations of these two datasets are introduced in Appendix~\ref{sec:appendix_detail_er} and Github. 
We present the statistics about these two datasets in Table~\ref{tab:exp_data_stat}. 
The document quantity for train/valid/test set of \textsc{Maven-Ere} and \textsc{OntoEvent} are respectively 2,913/710/857, and 2,622/747/746.

\emph{Baselines.} 
For trigger classification and event classification, we adopt models aggregated dynamic multi-pooling mechanism, \ie, DMCNN \cite{ACL2015_DMCNN} and DMBERT \cite{EMNLP2019_AD-DMBERT};  
sequence labeling models with conditional random field (CRF) \cite{ICML2001_CRF}, \ie, BiLSTM-CRF and BERT-CRF; 
generative ED models, \ie, TANL \cite{ICLR2021_TANL} and \textsc{Text2Event} \cite{ACL2021_Text2Event}. 
We also adopt some ED models considering document-level associations, \ie, MLBiNet \cite{ACL2021_MLBiNet} and CorED-BERT \cite{SIGIR2022_CorED}. 
Besides, we compare our energy-based hyperspheres with the vanilla hyperspherical prototype network (HPN) \cite{NIPS2019_HPN} and prototype-based model OntoED \cite{ACL2021_OntoED}. Note that unlike vanilla HPN \cite{NIPS2019_HPN} which represents all classes on one hypersphere, the HPN adopted in this paper represents each class with a distinct hypersphere. 
For event-relation extraction, we select RoBERTa \cite{arXiv2019_RoBERTa}, which is the same baseline used in \textsc{Maven-Ere} \cite{EMNLP2022_MAVEN-ERE}, and also serves as the backbone for most of recent ERE models \cite{ACL2022_ERE,AAAI2022_ERE}.

\subsection{Implementation Details} 
With regard to settings of the training process, Adam \cite{ICLR2014_Adam} optimizer is used, with the learning rate of 5e-5. 
The maximum length $L$ of a token sequence is 128, and the maximum quantity of event mentions in one document is set to 40 for \textsc{Maven-Ere} and 50 for \textsc{OntoEvent-Doc}. 
The loss ratios, $\mu_1$, $\mu_2$, $\mu_3$, for token, sentence and document-level energy function are all set to 1. 
The value of loss ratio, $\lambda_1$, $\lambda_2$, $\lambda_3$, for trigger classification, event classification and event-relation extraction depends on different tasks, and we introduce them in Appendix~\ref{sec:appendix_detail_imple}. 
We evaluate the performance of ED and ERE with micro precision (P), Recall (R) and F1 Score (F1).

\begin{table*}[!htbp]
\centering
\small
	\begin{tabular}{l | c c c | c c c}
	
	\toprule

	\multirow{3}*{\textbf{Model}} & \multicolumn{3}{c|}{\textbf{\textsc{Maven-Ere}}} & \multicolumn{3}{c}{\textbf{\textsc{OntoEvent-Doc}}} \\

	\cmidrule{2-7}

	& \textbf{P} & \textbf{R} & \textbf{F1} & \textbf{P} & \textbf{R} & \textbf{F1} \\

	\midrule

	DMCNN 
	& 61.74 $\pm$ 0.32 & 63.11 $\pm$ 0.34 & 62.42 $\pm$ 0.15 
	& 51.52 $\pm$ 0.87 & 52.84 $\pm$ 0.61 & 52.02 $\pm$ 0.36 \\


	DMBERT 
	& 59.45 $\pm$ 0.48 & \textbf{77.77} $\pm$ 0.21 & 67.39 $\pm$ 0.25 
	& 57.06 $\pm$ 1.04 & \textbf{72.97} $\pm$ 1.11 & \textbf{65.03} $\pm$ 0.45 \\




	HPN 
	& 62.80 $\pm$ 0.72 & 62.62 $\pm$ 0.99 & 62.71 $\pm$ 0.85 
	& \underline{61.18} $\pm$ 0.81 & 60.88 $\pm$ 0.79 & 61.03 $\pm$ 0.81 \\

	OntoED 
	& 67.82 $\pm$ 1.70 & 67.72 $\pm$ 1.52 & \underline{67.77} $\pm$ 1.61 
	& \textbf{64.32} $\pm$ 1.15 & 64.16 $\pm$ 1.31 & \underline{64.25} $\pm$ 1.22 \\

	TANL
	& \underline{68.73} $\pm$ 0.16 & 65.65 $\pm$ 0.63 & 67.15 $\pm$ 0.29 
	& 60.34 $\pm$ 0.71 & 62.52 $\pm$ 0.43 & 61.42 $\pm$ 0.51 \\

	\textsc{Text2Event} 
	& 61.14 $\pm$ 0.80 & 65.93 $\pm$ 0.69 & 63.44 $\pm$ 0.19
	& 56.76 $\pm$ 0.97 & \underline{66.78} $\pm$ 0.48 & 61.36 $\pm$ 0.77 \\


	\midrule

	\textbf{\textsc{\ours}} 
	& \textbf{72.91} $\pm$ 0.76 & \underline{72.81} $\pm$ 0.76 & \textbf{72.86} $\pm$ 0.77 
	& 58.92 $\pm$ 0.96 & 58.45 $\pm$ 1.08 & 58.69 $\pm$ 1.40 \\ 

	\quad w/o energy
	& 71.22 $\pm$ 0.58 & 71.07 $\pm$ 0.45 & 71.12 $\pm$ 0.45  
	& 56.12 $\pm$ 1.87 & 55.69 $\pm$ 1.66 & 55.91 $\pm$ 1.76  \\


	\bottomrule
	\end{tabular}
\vspace{-2mm}
	\caption{
	Performance (\%) of event classification on \textsc{Maven-Ere} \emph{valid set} and \textsc{OntoEvent-Doc} \emph{test set}. 
	\label{tab:exp_sent_ec}
	}
\vspace{-3.5mm}
\end{table*}




\subsection{Event Trigger Classification}

We present details of event trigger classification experiment settings in Appendix~\ref{sec:appendix_detail_imple_etc}. 
As seen from the results in Table~\ref{tab:exp_token_ed}, {\ours} demonstrates superior performance over all baselines, notably MLBiNet \cite{ACL2021_MLBiNet} and CorED-BERT \cite{SIGIR2022_CorED}, even if these two models consider cross-sentence semantic information or incorporate type-level and instance-level correlations. 
The main reason may be due to the energy-based nature of {\ours}. As seen from the last row of Table~\ref{tab:exp_token_ed}, the removal of energy functions from {\ours} can result in a performance decrease. Specifically for trigger classification, energy-based modeling enables capture long-range dependency of tokens and places no limits on the size of event structures. In addition, {\ours} also excels generative models, \ie, TANL \cite{ICLR2021_TANL} and \textsc{Text2Event} \cite{ACL2021_Text2Event}, thereby demonstrating the efficacy of energy-based modeling.

\subsection{Event Classification}

The specifics of event classification experiment settings are elaborated in Appendix~\ref{sec:appendix_detail_imple_ec}, with results illustrated in Table~\ref{tab:exp_sent_ec}. 
We can observe that {\ours} provides considerable advantages on \textsc{Maven-Ere}, while the performance on \textsc{OntoEvent-Doc} is not superior enough. 
\textsc{OntoEvent-Doc} contains overlapping where multiple event classes may exist in the same event mention, which could be the primary reason for {\ours} not performing well enough in this case. 
This impact could be exacerbated when joint training with other {\task} tasks. 
Upon comparison with prototype-based methods without energy-based modeling, \ie, HPN \cite{NIPS2019_HPN} and OntoED \cite{ACL2021_OntoED}, {\ours} is still dominant on \textsc{Maven-Ere}, despite HPN represents classes with hyperspheres and OntoED leverages hyperspheres integrated with event-relation semantics. If we exclude energy functions from {\ours}, performance will degrade, as seen from the last row in Table~\ref{tab:exp_sent_ec}. 
This insight suggests that energy functions contribute positively to event classification, which enable the model to directly capture complicated dependency between event mentions and event types, instead of implicitly inferring from data. Besides, {\ours} also outperforms generative models like TANL and \textsc{Text2Event} on \textsc{Maven-Ere}, indicating the superiority of energy-based hyperspherical modeling. 

\subsection{Event-Relation Extraction}
\label{sec:exp_ere}

\begin{table}[!htbp]
\vspace{2mm}
\centering
\small
\resizebox{\linewidth}{!}{
	\begin{tabular}{c l | c  | c }
	
	\toprule

	\multicolumn{2}{c|}{ \textbf{ERE Task} } 
	& \multicolumn{1}{c|}{ \textbf{RoBERTa} } & \multicolumn{1}{c}{ \textbf{\textsc{\ours}} } \\ 

	\midrule

	\multicolumn{1}{c|}{ \multirow{4}*{Temporal} }
	& \textsc{Maven-Ere}
	& \textbf{49.21} $\pm$ 0.33 
	& 39.64 $\pm$ 0.79 
	\\ \multicolumn{1}{c|}{}
	& \quad+joint 
	& \textbf{49.91} $\pm$ 0.58
	& 40.23 $\pm$ 0.34
	\\ \multicolumn{1}{c|}{}
	& \textsc{OntoEvent-Doc}
	& 37.68 $\pm$ 0.47 
	& \textbf{52.36} $\pm$ 0.71
	\\ \multicolumn{1}{c|}{}
	& \quad+joint 
	& 35.63 $\pm$ 0.70 
	& \textbf{65.69} $\pm$ 0.39 
	\\

	\midrule

	\multicolumn{1}{c|}{ \multirow{4}*{Causal} }
	& \textsc{Maven-Ere}
	& \textbf{29.91} $\pm$ 0.34 
	& 16.28 $\pm$ 0.53 
	\\ \multicolumn{1}{c|}{}
	& \quad+joint 
	& \textbf{29.03} $\pm$ 0.91
	& 16.31 $\pm$ 0.97 
	\\ \multicolumn{1}{c|}{}
	& \textsc{OntoEvent-Doc}
	& 35.48 $\pm$ 1.77 
	& \textbf{79.29} $\pm$ 2.15 
	\\ \multicolumn{1}{c|}{}
	& \quad+joint 
	& 44.99 $\pm$ 0.29
	& \textbf{67.76} $\pm$ 1.28 
	\\

	\midrule

	\multicolumn{1}{c|}{ \multirow{2}*{Subevent} }
	& \textsc{Maven-Ere}
	& 19.80 $\pm$ 0.44 
	& \textbf{19.91} $\pm$ 0.52 
	\\ \multicolumn{1}{c|}{}
	& \quad+joint 
	& 19.14 $\pm$ 2.81 
	& \textbf{21.96} $\pm$ 1.24
	\\

	\midrule

	\multicolumn{1}{c|}{ \multirow{2}*{All Joint} }
	& \textsc{Maven-Ere}
	& 34.79 $\pm$ 1.13
	& \textbf{37.85} $\pm$ 0.72
	\\ \multicolumn{1}{c|}{}
	& \textsc{OntoEvent-Doc}
	& 28.60 $\pm$ 0.13 
	& \textbf{54.19} $\pm$ 2.28 
	\\ 
	
	\bottomrule
	\end{tabular}
}
\vspace{-2mm}
	\caption{
	F1 (\%) performance of ERE on \textsc{Maven-Ere} \emph{valid set} and \textsc{OntoEvent-Doc} \emph{test set}. 
	‘‘+joint'' in the 2$_{nd}$ column denotes jointly training on all ERE tasks and evaluating on the specific one, with the same setting as \citet{EMNLP2022_MAVEN-ERE}. 
	‘‘All Joint'' in the last two rows denotes treating all ERE tasks as one task. 
	\label{tab:exp_ere}
	}
\vspace{-5mm}
\end{table}

We present the specifics of event-relation extraction experiment settings in Appendix~\ref{sec:appendix_detail_imple_ere}. 
As seen from the results in Table~\ref{tab:exp_ere}, {\ours} achieves different performance across the two ERE datasets. 
On \textsc{OntoEvent-Doc} dataset, {\ours} observably outperforms RoBERTa on all ERE subtasks, demonstrating the effectiveness of {\ours} equipped with energy-based hyperspheres, so that {\ours} can capture the dependency among event mention pairs and event-relation labels. 
While on \textsc{Maven-Ere}, {\ours} significantly outperforms RoBERTa on ERE subtasks referring to subevent relations or trained on all event-relations, but fails to exceed RoBERTa on ERE subtasks referring to temporal and causal relations. 
The possible reason is that \textsc{Maven-Ere} contains less positive event-relations than negative NA relations. 
Given that {\ours} models all these relations equivalently with the energy function, it becomes challenging to classify NA effectively. 
But this issue will be markedly improved if the quantity of positive event-relations decreases, since {\ours} performs better on subevent relations despite \textsc{Maven-Ere} having much less subevent relations than temporal and causal ones as shown in Table~\ref{tab:exp_data_stat}. 
Furthermore, even though \textsc{OntoEvent-Doc} containing fewer positive event-relations than NA overall, {\ours} still performs well. These results suggest that {\ours} excels in modeling classes with fewer samples.
Note that {\ours} also performs well when training on all event-relations (‘‘All Joint'') of the two datasets, indicating that {\ours} is still advantageous in the scenario with more classes.

\section{Further Analysis}
\label{sec:analysis}



\subsection{Analysis On Energy-Based Modeling}

\begin{figure}[!htbp] 
  \centering
  \includegraphics[width=0.8\linewidth]{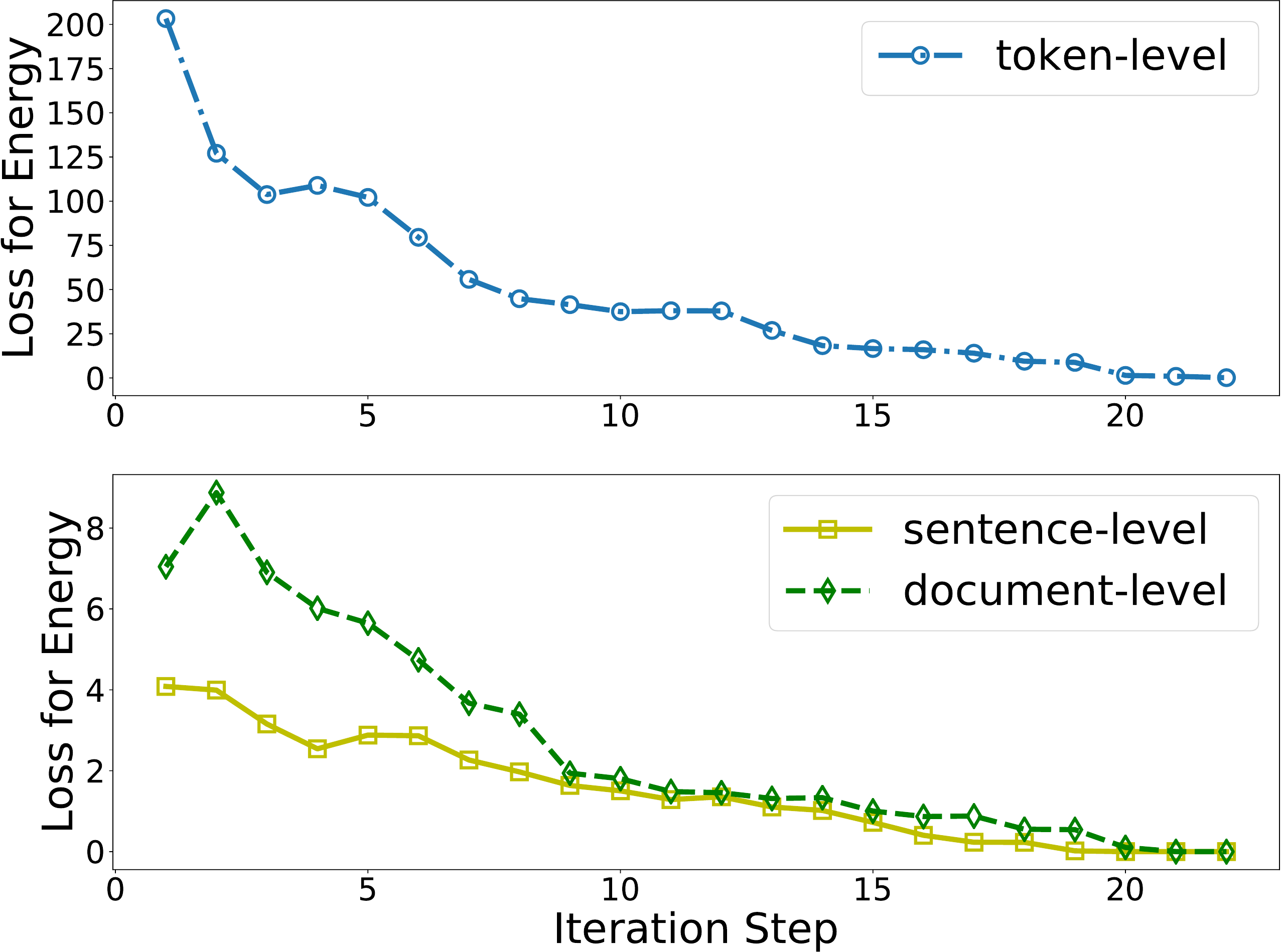}
  \vspace{-2mm}
  \caption{Illustration of loss for energy. 
  \label{fig:exp_loss_ene} }
  \vspace{-2mm}
\end{figure}

We list some values of energy loss defined in Eq~\eqref{eq:loss_token}, \eqref{eq:loss_sent} and \eqref{eq:loss_doc} when training respectively for token, sentence and document, as presented in Figure~\ref{fig:exp_loss_ene}. 
The values of token-level energy loss are observably larger than those at the sentence and document levels. This can be attributed to the fact that the energy loss is related to the quantity of samples, and a single document typically contains much more tokens than sentences or sentence pairs. 
All three levels of energy loss exhibit a gradual decrease over the course of training, indicating that \emph{{\ours}, through energy-based modeling, effectively minimizes the discrepancy between predicted results and ground truth.} 
The energy functions for token, sentence and document defined in Eq~\eqref{eq:token_energy}, \eqref{eq:sent_energy} and \eqref{eq:doc_energy}, reflect that \emph{the implementation of energy-based modeling in {\ours} is geared towards enhancing compatibility between input/output pairs.} 
The gradually-decreasing energy loss demonstrates that \emph{{\ours} can model intricate event structures at the token, sentence, and document levels through energy-based optimization, thereby improving the outcomes of structured prediction.} 

\subsection{Case Study: Energy-Based Hyperspheres}

\begin{figure}[!htbp] 
  \centering
  \includegraphics[width=0.74\linewidth]{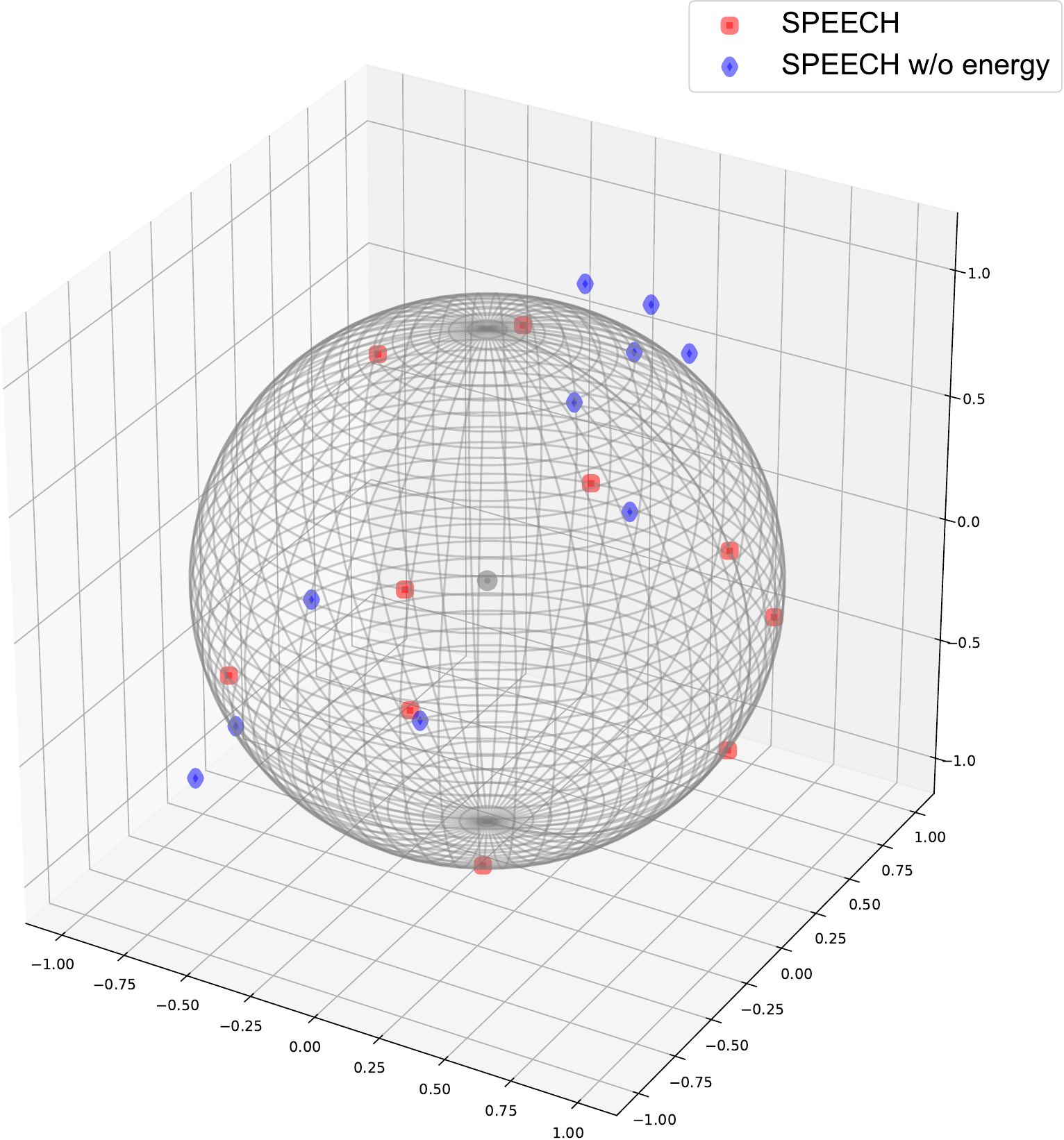}
  \caption{Visualization of an example event class. 
  \label{fig:exp_case} }
  \vspace{-1mm}
\end{figure}

As seen in Figure~\ref{fig:exp_case}, we visualize the event class embedding of ‘‘Attack'' and 20 event mention embeddings as generated by both {\ours} and {\ours} without energy functions. 
We observe that \emph{for {\ours} with energy-based modelling, the instances lie near the surface of the corresponding hypersphere, while they are more scattered when not equipped with energy-based modeling, 
which subsequently diminishes the performance of event classification.} 
This observation suggests that {\ours} derives significant benefits from modeling with energy-based hyperspheres. 
The visualization results further demonstrate the effectiveness of {\ours} equipped with energy-based modeling. 

\subsection{Error Analysis}
\label{sec:error_analysis}

We further conduct error analysis by a retrospection of  experimental results and datasets. 
(1) One typical error relates to the unbalanced data distribution. Considering every event type and event-relation contain different amount of instances, unified modeling with energy-based hyperspheres may not always be impactful. 
(2) The second error relates to the overlapping event mentions among event types, meaning that the same sentence may mention multiple event types. As \textsc{OntoEvent-Doc} contains many overlappings, it might be the reason for its mediocre performance on ED. 
(3) The third error relates to associations with event-centric structured prediction tasks. As trigger classification is closely related to event classification, wrong prediction of tokens will also influence classifying events.

\section{Conclusion and Future Work}
\label{sec:con_fw}

In this paper, we propose a novel approach entitled {\ours} to tackle event-centric structured prediction with energy-based hyperspheres. 
We represent event classes as hyperspheres with token, sentence and document-level energy, respectively for trigger classification, event classification and event relation extraction tasks. 
We evaluate {\ours} on two event-centric structured prediction datasets, and experimental results demonstrate that {\ours} is able to model manifold event structures with dependency and obtain effective event representations. 
In the future, we intend to enhance our work by modeling more complicated structures and extend it to other structured prediction tasks.

\clearpage

\section*{Acknowledgements}
We would like to express gratitude to the anonymous reviewers for their kind comments. 
This work was supported by the Zhejiang Provincial Natural Science Foundation of China (No. LGG22F030011), Yongjiang Talent Introduction Programme (2021A-156-G), CAAI-Huawei MindSpore Open Fund, Information Technology Center and State Key Lab of CAD \& CG, ZheJiang University, and NUS-NCS Joint Laboratory (A-0008542-00-00). 

\section*{Limitations}
\label{sec:limitations}

Although {\ours} performs well on event-centric structured prediction tasks in this paper, it still has some limitations. 
The first limitation relates to efficiency. As {\ours} involves many tasks and requires complex calculation, the training process is not very prompt. 
The second limitation relates to robustness. As seen in the experimental analysis in $\S$~\ref{sec:exp_ere}, {\ours} seems not always robust to unevenly-distributed data. 
The third limitation relates to universality. Not all event-centric structured prediction tasks can simultaneously achieve the best performance at the same settings of {\ours}. 

\balance
\bibliography{custom}
\bibliographystyle{acl_natbib}


\appendix

\section*{Appendices}
\label{sec:appendices}

\section{Multi-Faceted Event-Relations}
\label{sec:appendix_detail_er}

Note that \textsc{Maven-Ere} and \textsc{OntoEvent-Doc} both includes multi-faceted event-relations. 

\textsc{Maven-Ere} in this paper contains 6 temporal relations: BEFORE, OVERLAP, CONTAINS, SIMULTANEOUS, BEGINS-ON, ENDS-ON; 2 causal relations: CAUSE, PRECONDITION; and 1 subevent relation: subevent\_relations.

\textsc{OntoEvent-Doc} in this paper contains 3 temporal relations: BEFORE, AFTER, EQUAL; and 2 causal relations: CAUSE, CAUSEDBY. 

We also add a NA relation to signify no relation between the event mention pair for the two datasets. 

\section{Implementation Details for Different Tasks}
\label{sec:appendix_detail_imple}

\subsection{Event Trigger Classification}
\label{sec:appendix_detail_imple_etc}
\emph{Settings.} We follow the similar evaluation protocol of standard ED models \cite{ACL2015_DMCNN,SIGIR2022_CorED} on trigger classification tasks. 
We present the results in Table~\ref{tab:exp_token_ed} when jointly training with event classification and the whole ERE task (‘‘All Joint'' in Table~\ref{tab:exp_ere}). 
The backbone encoder is pretrained BERT \cite{NAACL2019_BERT}. 
The loss ratio, $\lambda_1$, $\lambda_2$, $\lambda_3$ in Eq~\eqref{eq:loss_all} are respectively set to 1, 0.1, 0.1 for both \textsc{OntoEvent-Doc} and \textsc{Maven-Ere}.

\subsection{Event Classification}
\label{sec:appendix_detail_imple_ec}
\emph{Settings.} We follow the similar evaluation protocol of standard ED models \cite{ACL2015_DMCNN,ACL2021_OntoED} on event classification tasks. 
We present the results in Table~\ref{tab:exp_sent_ec} when jointly training with trigger classification and all ERE subtasks (‘‘+joint'' in Table~\ref{tab:exp_ere}). 
The backbone encoder is pretrained DistilBERT \cite{NeurIPS2019-EMC2_DistilBERT}. 
The loss ratio, $\lambda_1$, $\lambda_2$, $\lambda_3$ in Eq~\eqref{eq:loss_all} are respectively set to 0.1, 1, 0.1 for \textsc{OntoEvent-Doc} and 1, 0.1, 0.1 for \textsc{Maven-Ere}.

\subsection{Event-Relation Extraction}
\label{sec:appendix_detail_imple_ere}
\emph{Settings.} We follow the similar ERE experiment settings with \citet{EMNLP2022_MAVEN-ERE} on several subtasks, by separately and jointly training on temporal, causal, and subevent event-relations. 
We present the results in Table~\ref{tab:exp_ere} when jointly training with trigger classification and event classification tasks. 
The backbone encoder is pretrained DistilBERT \cite{NeurIPS2019-EMC2_DistilBERT}. 
On \textsc{OntoEvent-Doc} dataset, the loss ratio, $\lambda_1$, $\lambda_2$, $\lambda_3$ in Eq~\eqref{eq:loss_all} are respectively set to 1, 0.1, 0.1 for all ERE subtasks. 
On \textsc{Maven-Ere} dataset, $\lambda_1$, $\lambda_2$, $\lambda_3$ are respectively set to 0.1, 0.1, 1 for ‘‘All Joint'' ERE subtasks in Table~\ref{tab:exp_ere}; 1, 1, 4 for ‘‘+joint''; 1, 0.1, 0.1 for ‘‘Temporal'' and ‘‘Causal''; and 1, 0.1, 0.08 for ‘‘Subevent''.

\end{document}